\documentclass[runningheads]{llncs}
\usepackage{graphicx}
\usepackage{amsmath,amssymb} 
\usepackage{color}

\usepackage{subfigure} 
\usepackage{wrapfig}
\usepackage{url}
\usepackage{amsmath}
\usepackage{amsfonts}
\usepackage{mathrsfs}
\usepackage{caption}

\usepackage[width=122mm,left=12mm,paperwidth=146mm,height=193mm,top=12mm,paperheight=217mm]{geometry}
\begin{document}

\newcommand\blfootnote[1]{%
  \begingroup
  \renewcommand\thefootnote{}\footnote{#1}%
  \addtocounter{footnote}{-1}%
  \endgroup
}

\newcommand{\fix}{\marginpar{FIX}}
\newcommand{\new}{\marginpar{NEW}}

\newcommand{\cX}{{\bf X}}
\newcommand{\cW}{{\bf \mathsf{W}}}
\newcommand{\cZ}{{\bf Z}}
\newcommand{\cQ}{{\bf Q}}
\newcommand{\sw}{{\bf w}}
\newcommand{\ow}{{\bf w}^{(out)}}
\newcommand{\cL}{\mathcal{L}}
\newcommand{\sL}{\ell}
\newcommand{\cg}{{\bf g}}
\newcommand{\sg}{{\bf \hat{g}}}
\newcommand{\cE}{{\mathbb{E}}}
\newcommand{\aP}{{\mathcal{P}}}
\newcommand{\aQ}{{\mathcal{Q}}}
\newcommand{\gp}{{\bf gp}}
\newcommand{\gq}{{\bf gq}}
\newcommand{\gf}{{\bf gf}}
\newcommand{\cWs}{{\bf \mathsf{W}}^{\star}}

\pagestyle{headings}
\mainmatter

\title{Top-down Learning for Structured Labeling with Convolutional Pseudoprior} 

\titlerunning{Top-down Learning for Structured Labeling with Convolutional Pseudoprior}

\authorrunning{Saining Xie, Xun Huang, Zhuowen Tu}

\author{Saining Xie$^{1\dag}$, Xun Huang$^{2\dag}$, Zhuowen Tu$^1$}


\institute{$^1$Dept. of CogSci and Dept. of  CSE, UC San Diego\\
	\email{ \{s9xie,ztu\}@ucsd.edu}\\
	$^2$Dept. of Computer Science, Cornell University\\
	\email{ xh258@cornell.edu}
}

\maketitle

\begin{abstract}
	Current practice in convolutional neural networks (CNN) remains largely bottom-up and the role of top-down process in CNN for pattern analysis and visual inference is not very clear.
		In this paper, we propose a new method for structured labeling by developing convolutional pseudoprior (ConvPP) on the ground-truth labels. Our method has several interesting properties: (1) compared with classic machine learning algorithms like CRFs and Structural SVM, ConvPP automatically learns rich convolutional kernels to capture both short- and long- range contexts; (2) compared with cascade classifiers like Auto-Context, ConvPP avoids the iterative steps of learning a series of discriminative classifiers and automatically learns contextual configurations; (3) compared with recent efforts combining CNN models with CRFs and RNNs, ConvPP learns convolution in the labeling space with improved modeling capability and less manual specification; (4) compared with Bayesian models like MRFs, ConvPP capitalizes on the rich representation power of convolution by automatically learning priors built on convolutional filters. We accomplish our task using pseudo-likelihood approximation to the prior
		under a novel fixed-point network structure that facilitates an end-to-end learning process. We show state-of-the-art results on sequential labeling and image labeling benchmarks.\blfootnote{$\dag$ equal contribution.}
		
\end{abstract}

	\section{Introduction}
	
	Structured labeling is a key machine learning problem: structured inputs and 	outputs are common in a wide range of machine learning and computer vision applications \cite{elman1990finding,lafferty2001conditional,shotton2006textonboost}. 
	The goal of structured labeling is to simultaneously assign labels (from some fixed label set) to individual elements in a structured input. 
	Markov random fields (MRFs) \cite{geman1984stochastic} and conditional random fields (CRFs)
	\cite{lafferty2001conditional} have been widely used to model the correlations between 
	the structured labels. However, due to the heavy computational burden in their training and testing/inference stages, MRFs and CRFs are often limited to 
	capturing a few neighborhood interactions with consequent restrictions of their modeling 
	capabilities. Structural SVM methods \cite{tsochantaridis05largemargin} and maximum margin Markov 
	networks ($\textrm{M}^{3}\textrm{N}$) \cite{taskar2003max} capture correlations 
	in a way similar to CRFs, but they try to specifically maximize the prediction 
	margin; these approaches are likewise limited in the range of contexts, again due 
	to associated high computational costs. When long range contexts are used, 
	approximations are typically used to trade between accuracy and efficiency \cite{finley2008training}.
	Other approaches to capture output variable dependencies have been proposed by introducing
	classifier cascades.  For example, cascade models \cite{tu2008auto,heitz2009cascaded,daume2009search} in the spirit of stacking \cite{wolpert1992stacked}, are proposed to take the outputs 
	of classifiers of the current layer as additional features for the next classifiers in 
	the cascade. Since these approaches perform direct label prediction (in the form 
	of functions) instead of inference as in MRFs or CRFs, the cascade models 
	\cite{tu2008auto,heitz2009cascaded} are able to model complex and long-range contexts.
	
	Despite the efforts in algorithmic development with very encouraging results produced in the past, the problem of structured labeling remains a challenge. To capture high-order configurations of the interacting labels, top-down information, or prior offers assistance in both training and testing/inference. The demonstrated role of top-down information in human perception \cite{ames1951visual,marr1982vision,gibson2002theory} provides a suggestive indication of the form that top-down information could play in structured visual inference. Systems trying to explicitly incorporate top-down information under the Bayesian formulation point to a promising direction \cite{kersten2004object,tu2005image,borenstein2008combined,wu2011topdown} but in the absence of a clear solution.  Conditional random fields family models that learn the posterior directly \cite{lafferty2001conditional,tu2008auto,heitz2009cascaded,krahenbuhl2011efficient} alleviates some burdens on learning the labeling configuration, but still with many limitations and constraints. The main difficulty is contributed by the level of complexity in building high-order statistics to capture a large number of interacting components within both short- and long- range contexts.
	
	From a different angle, building convolutional neural networks for structured labeling \cite{long2015fully} has resulted in systems that greatly outperform many previous algorithms. Recent efforts in combining CNN with CRF and RNN models \cite{zheng2015conditional,lin2015deeply} have also shed light onto the solution of extending CNN to structured prediction. However, these approaches still rely on CRF-like graph structure with limited neighborhood connections and heavy manual specification. More importantly, the explosive development in modeling data using layers of convolution has not been successfully echoed in modeling the prior in the label space. 
	
	In this paper, we propose a new structured labeling method by developing convolutional pseudoprior (ConvPP) on the ground-truth labels, which is infeasible by directly learning convolutional kernels using the existing CNN structure. We accomplish our task by developing a novel end-to-end fixed-point network structure using pseudo-likelihood approximation \cite{besag1977efficiency} to the prior that learns convolutional kernels and captures both the short- and the long- range contextual labeling information. We show state-of-the-art results on benchmark datasets in sequential labeling and popular image labeling.
	
	\section{Related Work}

	We first summarize the properties of our proposed convolutional pseudoprior (ConvPP) method: (1) compared with classical machine learning algorithms like CRFs \cite{lafferty2001conditional}, Structural SVM (\cite{tsochantaridis05largemargin}), and max-margin Markov networks \cite{taskar2003max} , ConvPP automatically learns rich convolutional kernels to capture both the short- and the long- range contexts. (2) Compared with cascade classifiers \cite{tu2008auto,heitz2009cascaded}, ConvPP avoids the time-consuming steps of iteratively learning a series of discriminative classifiers and it automatically learns the contextual configurations (we have tried to train a naive auto-context type of fully convolutional model instead of modeling prior directly from the ground-truth label space but without much success; the overall test error did not decrease after long-time training with many attempts of parameter tweaking; this is possibly due to the difficulty in capturing meaningful contexts on the predicted labels, which are noisy). (3) Compared with recent efforts combining CNN models with CRFs and RNNs \cite{zheng2015conditional,lin2015deeply}, ConvPP learns convolution in the labeling space with improved modeling capability and less manual specification. (4) Compared with Bayesian models \cite{tu2005image,zhu2006stochastic} ConvPP capitalizes on the rich representation power of CNN by automatically learning convolutional filters on the prior.

	In addition, we will discuss some other related work. \cite{he2004multiscale} addresses structured (image) labeling tasks by building a multi-scale CRF with handcrafted features and constrained context range, whereas in our work we learn the context automatically in an end-to-end network. \cite{kae2013augmenting} also combines RBM and CRF to incorporate shape prior for face segmentation. \cite{krahenbuhl2011efficient} is able to learn a large neighborhood graph but under a simplified model assumption; in \cite{henaff2015deep} deep convolutional networks are learned on a graph but the focus there is not for structured labeling; deep belief nets (DBN) \cite{hinton2006fast} and auto-encoders \cite{snoek2012nonparametric} are generative models that potentially can be adapted for learning the prior but it a clear path for structured labeling is lacking.
	Our work is also related to recurrent neural networks (RNNs) \cite{elman1990finding}, but ConvPP has its particular advantage in: (1) modeling capability as explicit convolutional features are learned on the label space; (2) reduced training complexity as the time-consuming steps of computing recurrent responses are avoided by directly using the ground truth labels as a fixed-point model. In deep generative stochastic  networks\cite{bengio2013deep}, pseudo-likelihood is used to train a deep generative model, but not for learning priors with CNN.
	
	To summarize, ConvPP builds an end-to-end system by learning a novel hybrid model with convolutional pseudopriors on the labeling space and traditional bottom-up convolutional neural networks for the appearance.
	
	
	\section{Formulations}
	
	\begin{figure*}[!htp]
		\begin{center}
			\includegraphics[width=1\textwidth]{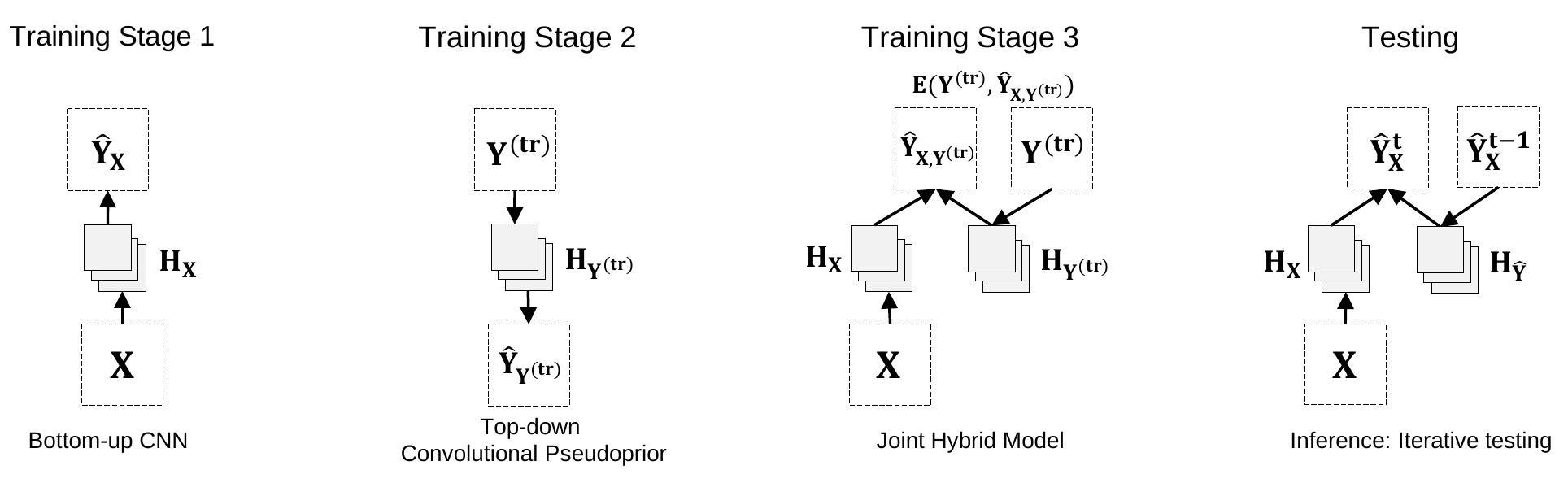} 
			\caption{\footnotesize The architecture of our ConvPP framework. At the first training stage, we train a bottom-up CNN model with image data as input; at the second training stage, we train a top-down convolutional pseudoprior model from ground-truth label maps. The hidden representations are then concatenated and the network is fine-tuned with the joint hybrid model. At inference, since we don't have the ground-truth label anymore, we iteratively feed predictions to the convolutional pseudoprior part. }
			\label{fig:Pipeline}
		\end{center}
	\end{figure*}
	We first briefly discuss the structured labeling problem and understand it from the Bayesian point of view. Let $\mathcal{X}$ be the space of input observations and $\mathcal{Y}$ be the space of possible labels. Assume any data-label pairs $(\mathbf{X},\mathbf{Y})$ follow a joint distribution $p(\mathbf{X},\mathbf{Y})$. We seek to learn a mapping $F: \mathcal{X} \rightarrow \mathcal{Y}$ that minimizes the expected loss. For a new input sample $X\in\mathcal{X}$, we want to determine the optimal labeling $\mathbf{Y}^*$ that maximizes the posterior probability $p(\mathbf{Y}|\mathbf{X})$. 
	\begin{equation}
		\mathbf{Y}^* = \arg \max_\mathbf{Y} p(\mathbf{Y}|\mathbf{X}) = \arg \max_\mathbf{Y} p(\mathbf{X}|\mathbf{Y})p(\mathbf{Y})
		\label{eq:Bayes}
	\end{equation}
	In the scenario of structured labeling such as pixel-wise labeling, intuitively, the labeling decision should be made optimally by considering both the appearance and the prior, based on the Bayes rule in Eq. (\ref{eq:Bayes}). However, learning both $p(\mathbf{X}|\mathbf{Y})$ and $p(\mathbf{Y})$ for complex structures is considered as very challenging.
	Our motivation here is to capitalize on the rich representational and compelling computational power of CNN for modeling both the appearance and prior. A large amount of work in the past using CNN has been primarily focused on training strong classifiers for predicting semantic labels (a discriminative way of modeling the appearance, \cite{long2015fully}), but rarely on the prior part (top-down information).
	
	To formulate our structured labeling problem, here we consider a graph, $\mathcal{G} = (\mathcal{V}, \mathcal{E})$. In a 1-D sequential labeling case, the graph is equivalent to a chain. The edge set $\mathcal{E}$ decides the graph topology, and hence the neighborhoods of every node. We denote $\mathcal{N}_i/i$  as the neighborhoods of node $v_i$. For each node $v_i$, we have its associated  data $\mathbf{x}_i$, ground-truth label $y_i$, and ground-truth labels for all the neighborhoods of $v_i$ as $\mathbf{y}_{\mathcal{N}_i/i}$. 
	Inspired by pseudo-likelihood \cite{besag1977efficiency} and the hybrid model in \cite{tu2008brain}, we make an approximation to the posterior in Eq. (\ref{eq:Bayes}) as follows:
	\begin{equation}
		p(\mathbf{Y}|\mathbf{X})  \propto 		p(\mathbf{Y}) p(\mathbf{X}|\mathbf{Y})  
		\dot{\propto}\ \prod_{i} p(y_i|\mathbf{Y}_{\mathcal{N}_i/i}) \cdot \prod_{i} p(y_i|\mathbf{X})
		\label{eqn2.qo}
	\end{equation}
	where $\mathbf{Y}_{\mathcal{N}_i/i}$ encodes a neighborhood structure (contexts) of $y_i$ for computing a pseudo-likelihood $p(y_i|\mathbf{Y}_{\mathcal{N}_i/i})$ \cite{besag1977efficiency} to approximate $p(\mathbf{Y})$, {\em but now as a prior}.
		
In addition, to see how the approximation to $p(\mathbf{X}|\mathbf{Y})$ by $\prod_{i} p(y_i|\mathbf{X})$ is obtained from Bayesian to conditional probability:	(1) assume pseudo-likelihood on each pixel i to approximate the likelihood term $p(\mathbf{X}|\mathbf{Y})$, using $p(\mathbf{x}_i | x_{\mathcal{N}_i/i},Y)$. Note that $\mathbf{x}_{\mathcal{N}_i/i}$ includes all the neighboring pixels of pixel i but excluding i;
(2) assume independence, to approximate $p(\mathbf{x}_i | \mathbf{x}_{\mathcal{N}_i/i},Y)$  by $p(\mathbf{x}_i | \mathbf{x}_{\mathcal{N}_i/i},y_i)$;
(3) $p(\mathbf{x}_i | \mathbf{x}_{\mathcal{N}_i/i},y_i)=p(\mathbf{x}_i ,y_i| \mathbf{x}_{\mathcal{N}_i/i})/P(y_i|\mathbf{x}_{\mathcal{N}_i/i}))$ and drop $p(y_i|\mathbf{x}_{\mathcal{N}_i/i})$ for another approximation. This leads to
$p(\mathbf{x}_i ,y_i| \mathbf{x}_{\mathcal{N}_i/i})$ which is 
$p(y_i| \mathbf{x}_{\mathcal{N}_i/i},\mathbf{x}_i )p(\mathbf{x}_i|\mathbf{x}_{\mathcal{N}_i/i})$;
(4) the above becomes $p(y_i|\mathbf{x}_{\mathcal{N}_i}) = p(y_i|\mathbf{X})$ when dropping $p(\mathbf{x}_i|\mathbf{x}_{\mathcal{N}_i/i})$.
	
	This hybrid model is of special interest to us since: (1) our end-to-end deep learning framework allows a discriminative convolutional neural network (CNN) to be trained to compute $p(y_i|\mathbf{X})$ to model the appearance; (2) by directly working on the ground-truth labels $\mathbf{Y}$, we also learn a convolutional pseudoprior as $\prod_{i} p(y_i|\mathbf{Y}_{\mathcal{N}_i/i})$ using a pseudo-likelihood approximation.

	Given a training data pair $p(\mathbf{X},\mathbf{Y}^{(tr)})$, to solve an approximated MAP problem with convolutional pseudoprior,
	\begin{align}
		\label{eqn:convpp}
		\mathbf{Y}^{(tr)} = \arg \max_{\mathbf{Y}} \prod_i p(y_i|\mathbf{Y}_{\mathcal{N}_i/i}; \mathbf{w}_2) \cdot \prod_{i} p(y_i|\mathbf{X}; \mathbf{w}_1)
	\end{align}
	
	From another perspective, the above learning/inference scheme can be motivated by the fixed-point model \cite{li2013fixed}.  Denote $\mathbf{Q}$ as the one-hot encoding of labeling $\mathbf{Y}$, and therefore $\mathbf{Q}^{(tr)}$ as the one-hot encoding of ground-truth training labeling $\mathbf{Y}^{(tr)}$. The fixed-point model solve the problem with the formulation for a prediction function $\mathbf{f}$, 
	\begin{align}
		\mathbf{Q} = \mathbf{f}(\mathbf{x}_1,
		\mathbf{x}_2,\cdots, \mathbf{x}_n, \mathbf{Q}; \sw)
	\end{align}
	where {$\mathbf{f}(\cdot) =
		[f(\mathbf{x}_1,\mathbf{Q}_{\mathcal{N}_1};
		\sw),\cdots, f(\mathbf{x}_n,\mathbf{Q}_{\mathcal{N}_n}; \sw)]^T$}, $\mathbf{Q} = [\textbf{q}_1,\textbf{q}_2, \cdots, \textbf{q}_n]^T$, and $\textbf{q}_i = f(\mathbf{x}_i, \mathbf{Q}_{\mathcal{N}_i})$. $\sw=(\sw_1,\sw_2)$. To get the labeling of a structured input graph $\mathcal{G}$, one can
	solve the non-linear system of equations $\mathbf{Q} =
	\mathbf{f}(\mathbf{x}_1, \mathbf{x}_2,\cdots,
	\mathbf{x}_n,\mathbf{Q}; \sw)$, which is generally a very
	difficult task. However, \cite{li2013fixed} shows that in many cases we can assume $\mathbf{f}$ represents so called contraction mappings, and so have an attractive fixed-point (a ``stable state'') for each structured input. When using the ground-truth labeling in the training process, that ground-truth labeling $\mathbf{Q}^{(tr)}$ is assumed to be the stable state: $\mathbf{Q}^{(tr)} =
	\mathbf{f}(\mathbf{x}_1, \mathbf{x}_2,\cdots, \mathbf{x}_n,\mathbf{Q}^{(tr)};\sw)$. 
	
	Next, we discuss the specific network architecture design and our training procedure.
	The pipeline of our framework is shown in Figure~\ref{fig:Pipeline} consisting of three stages: (1) training $\sw_1$
	for $p(y_i|\mathbf{X}; \sw_1)$; (2) training $\sw_2$ for $p(y_i|\mathbf{Y}_{\mathcal{N}_i/i}; \mathbf{w}_2)$; and (3) fine-tuning for $\prod_i p(y_i|\mathbf{Y}_{\mathcal{N}_i/i}; \mathbf{w}_2) \cdot p(y_i|\mathbf{X};\sw_1)$ jointly.
	
	At the first stage, we independently train a standard bottom-up CNN on the input data, in our work, we are especially interested in end-to-end architectures such as FCN \cite{long2015fully}. Without loss of generality, we abstractly let the feature representations learned by FCN be $\mathbf{H_X}$, and network predictions be $\mathbf{\hat{Y}_X}$, The error is computed with respect to the ground-truth label $\mathbf{Y}^{(tr)}$ and back-propagated during training.
	Similarly, at the second stage, we train a convolutional pseudoprior network on the ground-truth label space. Conceptually the prior modeling is a top-down process. Implementation-wise, the ConvPP network is still a CNN. However, the most notable difference compared with a traditional CNN is that, the ground-truth labels are not only used as the supervision for back-propagation, but also used as the network input. We learn hidden representation $\mathbf{H_{Y^{(tr)}}}$ and aim to combine this with the hierarchical representation $\mathbf{H_X}$ learned in the bottom-up CNN model. 
	Thus, combining pre-trained bottom-up CNN network and top-down ConvPP network, we build a joint hybrid model network in the third training stage. We concatenate $\mathbf{H_X}$ and $\mathbf{H_Y}$ (which can be finetuned) and learn a new classifier on top to produce the prediction $\mathbf{\hat{Y}_{X,Y^{(tr)}}}$. The joint network is still trained with back-propagation in an end-to-end fashion.
	
	At inference time, since we do not have the ground-truth label $\mathbf{Y}^{(tr)}$ available anymore, we follow the fixed-point motivation discussed above. We iteratively feed predictions $\mathbf{\hat{Y}_{X}^{t-1}}$ made at previous iteration, to the ConvPP part of the hybrid model network. The starting point $\mathbf{\hat{Y}_{X}^{0}} = \mathbf{0}$ can be a zero-initialized dummy prediction, or we can simply use $\mathbf{\hat{Y}_{X}^{0}} = \mathbf{\hat{Y}_{X}}$ given the pre-trained bottom-up CNN model.
	
	This conceptually simple approach to approximate and model the prior naturally faces two challenges: 1) How do we avoid trivial solutions and make sure the ConvPP network can learn meaningful structures instead of converging to an identity function? 2) When the bottom-up CNN is deep and involves multiple pooling layers, how to match the spatial configurations and make sure that, $\mathbf{H_X}$ and $\mathbf{H_{Y^{(tr)}}}$ are compatible in terms of the appearances and structures they learn.
	
\begin{figure}[!htp]
			\begin{center}
				\includegraphics[width=0.6\textwidth]{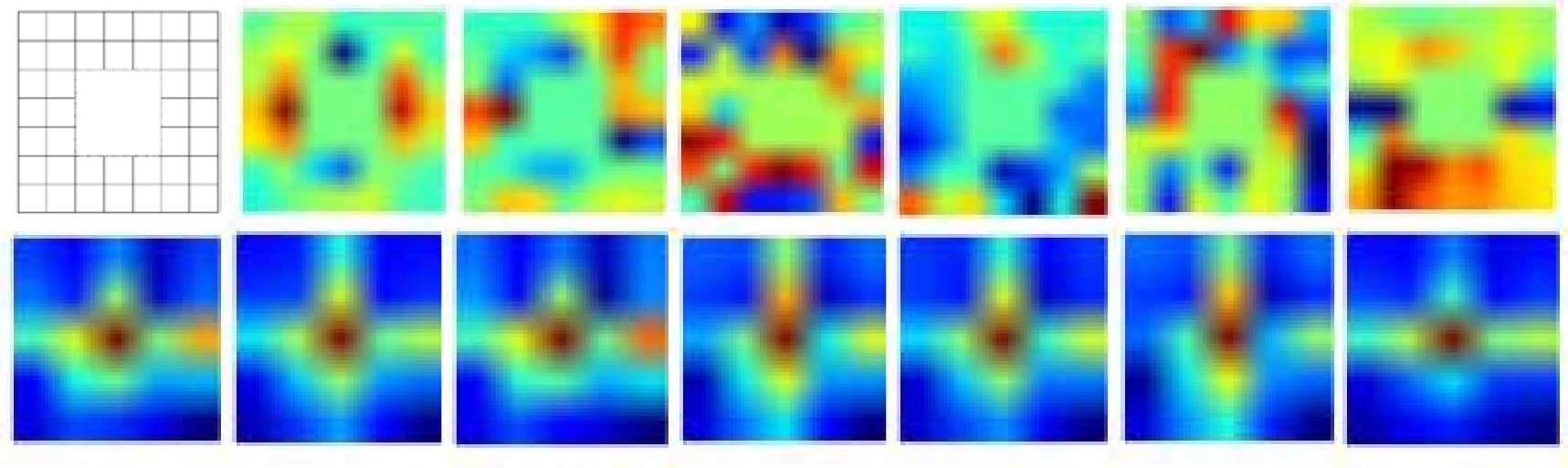} 
				\caption{\footnotesize The first row shows the donut filter we use for training the ConvPP network, and examples of learned real-valued filters after the ConvPP network training, originally initialized randomly. Note that a square hole (showing in light-green color) in the center is kept zero-valued during training, and it enforces the filter to learn non-trivial transformations. The second row shows trivial filters (close to identity transformation) are learned by a conventional auto-encoder. }
				\label{fig:Donut}
			\end{center}
		\end{figure}

	\textbf{ConvPP network architectures.} We will now explain the architecture design in ConvPP network to address the issues above. We have explored possible ways to avoid learning a trivial solution. Besides the ConvPP architecture design, one might think learning a convolutional auto-encoder on the ground-truth label space can achieve similar goal. However, we found that when training an auto-encoder on label space, the problem of trivial recovery is even more severe compared to training auto-encoders on natural images. We tried different regularization and sparsification techniques presented in recent convolutional auto-encoder works (e.g. \cite{makhzani2015winner}), but none of them work in our case. See Figure~\ref{fig:Donut} for a visual comparison. We conjecture that the reasons could be (1) the ground-truth labels are much simpler in their appearances, compared with natural images with rich details. Thus the burden of being identically reconstructed is greatly eased; (2) on the other hand, the structures like class inter-dependencies, shape context and relative spatial configurations are highly complex and subtle, make it really challenging to learn useful representations.
	
		\textbf{Donut filter.} Here we use a very simple yet effective approach to conquer the issues: our ConvPP network contains only a single convolutional layer, where we apply filters referred as ``donut filters''. The name comes from the way we modify the traditional convolution kernels: we make a hole in the center of the kernel. Figure~\ref{fig:Donut} shows an example where a 3 by 3 hole is in the middle of a 7 by 7 convolution filter. Given that we only have one convolutional layer, we impose a hard constraint on the ConvPP representation learning process: the reconstruction of the central pixel label will never see its original value, instead it can only be inferred from the neighboring labels. This is aligned with our pseudoprior formulation in Eq.~(\ref{eqn:convpp}).
	
	Donut filters are not supposed to be stacked to form a deep variant, since the central pixel label information, even though cropped from one layer, can be propagated from lower layers which enables the network to learn a trivial solution. Empirically we found that one hidden convolution layer with multiple filters is sufficient to approximate and model the useful prior in the label space.

	\textbf{Multi-scale ConvPP.} A natural question then becomes, since there is only one convolution layer in the ConvPP network, the receptive field size is effectively the donut filter kernel size. Small kernel size will result in very limited range of context that can be captured, while large kernel size will make the learning process extremely hard and often lead to very poor local minimum. 
		
			\begin{figure}[!htp]
			\begin{centering}
				\includegraphics[width=0.65\linewidth]{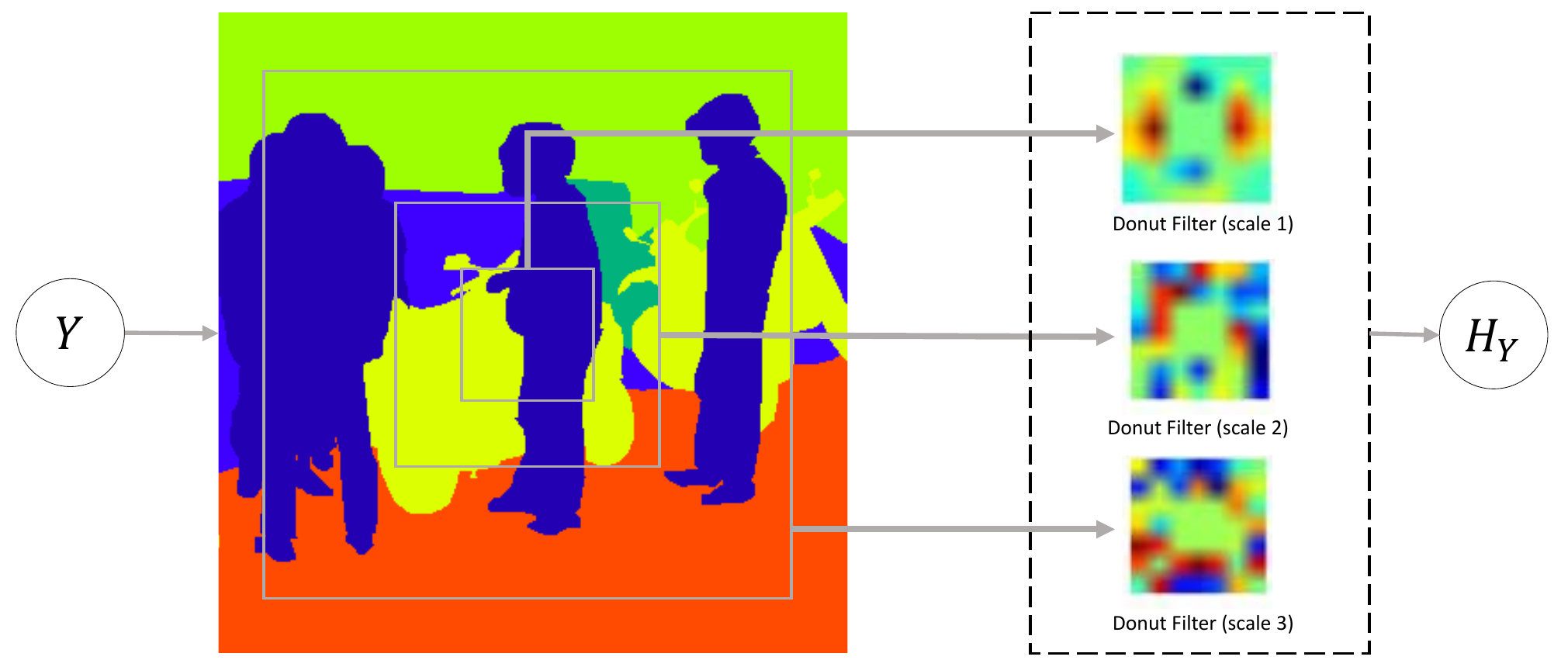} 
				\caption{\footnotesize Multiple donut filter layers with the same kernel size are integrated into different depth of the joint network. Multi-scale context learning is naturally handled. }
				\label{fig:multiscale}
				\par\end{centering}
		\end{figure}
		
	
	To combat this issue, as illustrated in Figure~\ref{fig:multiscale}, we leverage the unique simplicity of ground-truth labeling maps, and directly downsample the input ground-truth maps through multiple pooling layers to form a chain of multi-scale ground-truth maps. The in-network pooling layers also keep the network to be end-to-end trainable. This enables us to freely learn the ConvPP representations on different scales of the ground-truth label space. The flexibility of the useful context that ConvPP can capture, now comes from two aspects: (1) the convolution filter weights can be automatically learned during training. (2) the context range is also handled explicitly by multi-scale architecture design. One can imagine that ConvPP representations, learned on low-resolution ground-truth maps, are capable of modeling complex long range and high order semantic context, global object shape and spatial configuration, whereas representations learned on high-resolutions ground-truth maps are supposed to model local structures like local smoothness.
	
	Given that we can learn $\mathbf{H_Y}$ from different scales, we are readily able to build the spatial correspondences between $\mathbf{H_X}$ and  $\mathbf{H_Y^{(tr)}}$. One can concatenate the $\mathbf{H_Y}$ to any convolutional feature maps learned in the bottom-up CNN network, as long as they passed through the same number of downsampling layers.	
	
	Because our convolutional pseudoprior is learned directly from the ground-truth label space, and it does not condition on the input data at all, the choice of bottom-up CNN models are flexible. The complementary structural information provided by the ConvPP allows us to easily improve on state-of-the-art CNN architectures such as Fully Convolutional Neural Networks (FCN). 
	
	\section{Experiments}
		In this section, we show experimental results on four benchmark datasets across different domains, namely FAQ (Natural language processing), OCR (Sequential image recognition), Pascal-Context (Semantic segmentation) and SIFT Flow (Scene labeling).
	
	\subsection{Sequential Labeling: 1-D case}
		
	\begin{table}[t]
		\center
		\caption{\footnotesize An experimental comparison on the OCR dataset by varying the number of training data. It demonstrates that the generalization error monotonically decreases when adding more data.
		}
		\label{OCR_gen}
		\begin{tabular}{@{}ccllllllll@{}}
			\hline
			\hline
			
			Training Data Percentage (\%) & 10 & 20   & 30   & 40   & 50   & 60   & 70   & 80   & 90 \\ \hline
			Generalization Error(\%)                                            & \textbf{6.49}                       & \textbf{3.28} & \textbf{2.09} & \textbf{1.67} & \textbf{1.55} & \textbf{1.03} & \textbf{0.92} & \textbf{0.72} & \textbf{0.57}  \\ \hline
		\end{tabular}
	\end{table}
	
	First, we explore the effectiveness of our proposed framework on two 1-D structured (sequential) labeling tasks: handwritten OCR \cite{taskar2003max} and FAQ sentence labeling \cite{mccallum2000maximum}. In these two 1-D toy examples, the pseudoprior model is implemented as a fully connected layer whose inputs are one-hot-encoding of neighboring labels (excluding the to-be-predicted label itself). When we slide the model over the input sequences, this layer can be viewed as a 1D convolutional layer with kernel size m (m is the context window size) and hole size 1.
	
	\textbf{Handwritten OCR} This dataset contains $6,877$ handwritten words, corresponding to various writings of $55$ unique words. Each word is represented as a series of handwritten characters; there are $52,152$ total characters. Each character is a binary $16\times8$ image, leading to 128-dimensional binary feature vectors. Each character is one of the 26 letters in the English alphabet. The task is to predict the identity of each character. We first resize all the OCR characters 
	to the same size ($28\times28$) and build a standard 5-layer LeNet \cite{lecun1998gradientbased}. The label context part has a single-hidden-layer MLP with 100 units. We normalize each image to zero-mean and unit-variance.
	
	\begin{figure}
		\hfill
		\subfigure[]{\includegraphics[width=5cm]{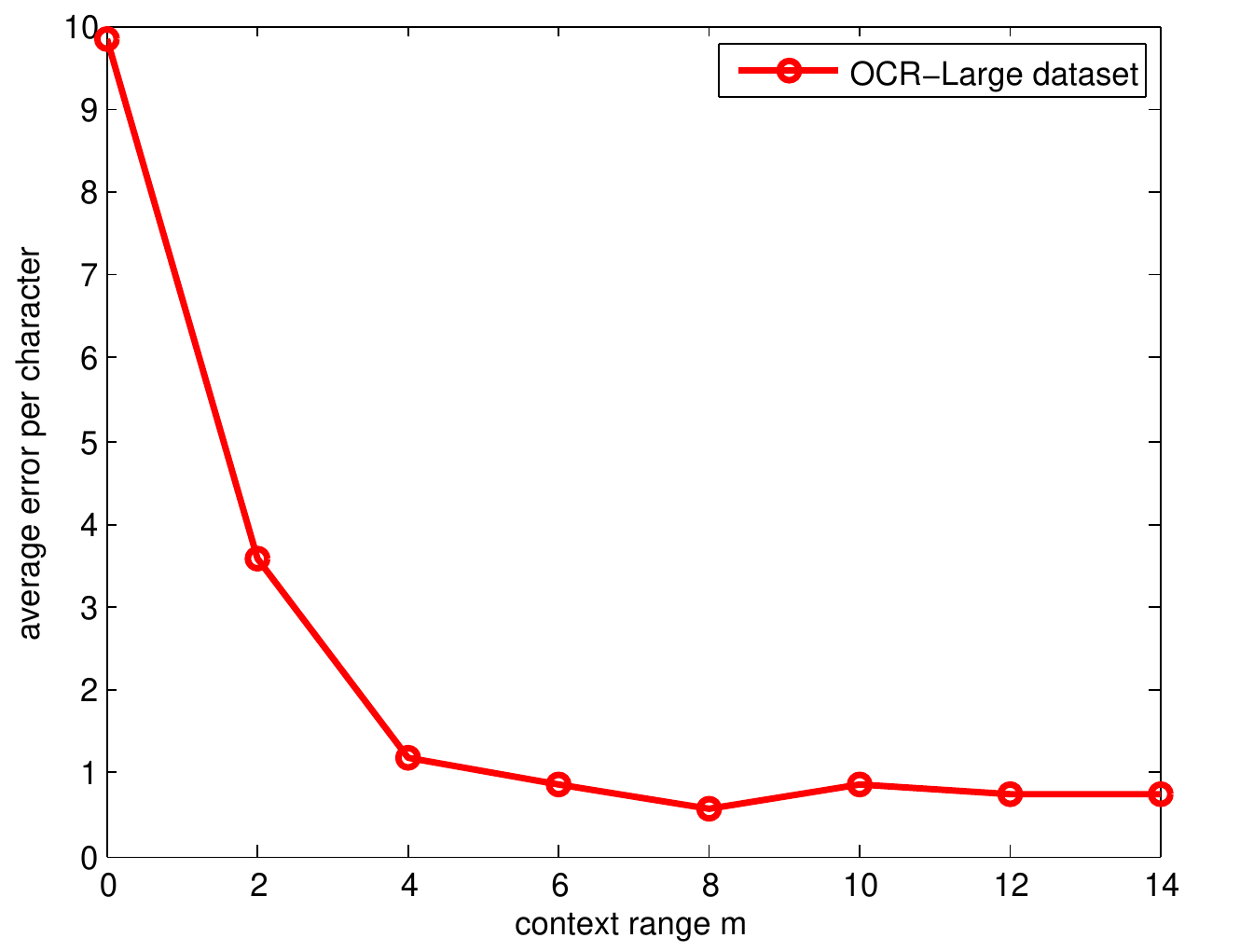}}
		\hfill
		\subfigure[]{\includegraphics[width=5cm]{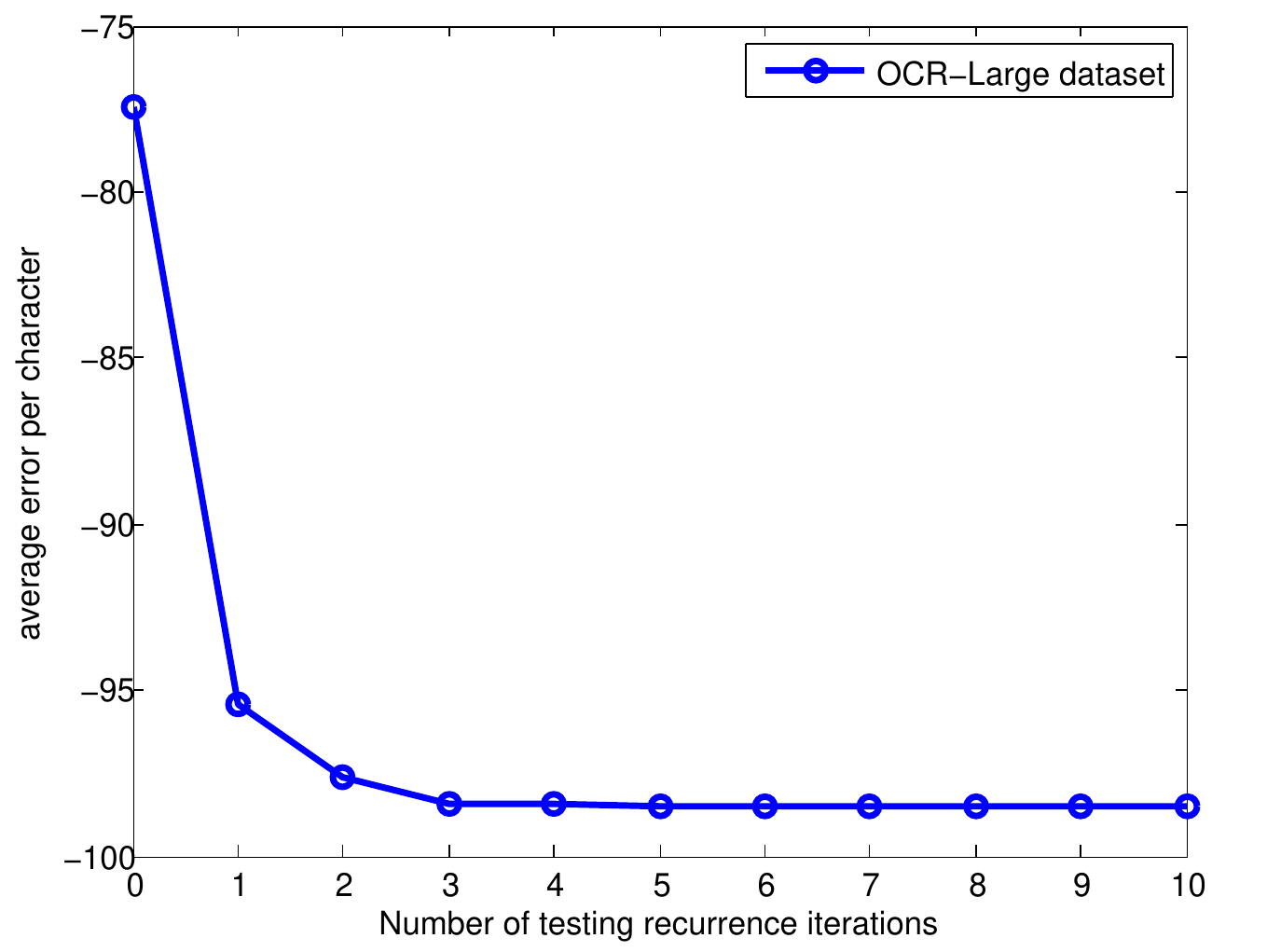}}
		\hfill
		\caption{\footnotesize (a) comparison of the generalization error on the OCR handwritten dataset by varying the context window length. (b) the generalization error on the OCR handwritten dataset as the number of testing iterations varies. }
		\label{fig:1d_curve}
	\end{figure}
	
	\textbf{FAQ} The FAQ dataset consists of 48 files collecting questions and answers gathered from 7 multi-part UseNet FAQs. There are a total of $55,480$ sentences across the 48 files. Each sentence is represented with 24-dimensional binary feature vector. \cite{mccallum2000maximum} provides a description of the features). We extended the feature set with all pairwise products of the original 24 features, leading to a $600$-dimensional feature representation. Each sentence in the FAQ dataset is given one of four labels: (1) head, (2) question, (3) answer, or (4) tail. The task is to predict the label for each sentence. We train a 3-hidden layer fully-connected network with [32, 64, 128] hidden units respectively. A single-hidden-layer MLP with 100 hidden units is trained on ground-truth labels.
	
	For both of the dataset, two hyper-parameters are specified by cross-validation: we set the context window size to be 7 (for OCR) and 5 (for FAQ); the number of iterations during testing to be 10.

	%
	%
	%
	%
	
	\begin{table}[!htp]
		\parbox[t][][t]{.45\linewidth}{
			\centering
						\caption{\footnotesize Performance (error rate (\%)) of structured labeling methods on the OCR dataset.}
						\label{res:OCR}
			\begin{tabular}{l|l|l}
				\hline 
				Methods &small		&large		\tabularnewline
				\hline 
				Linear-chain CRF \cite{do2010neural} 			&21.62	& 14.20	\tabularnewline
				$\textrm{M}^{3}\textrm{N}$ \cite{do2010neural} 		&21.13	& 13.46	\tabularnewline
				\textsc{Searn} \cite{daume2009search}	 		&	-	& 9.09	\tabularnewline
				SVM + CRF (\cite{hoefel2008learning}			&	-	& 5.76	\tabularnewline
				Neural CRF \cite{do2010neural} 			&10.8		& 4.44	\tabularnewline
				\scriptsize{Hidden-unit CRF \cite{vandermaaten2011hidden}}			&	18.36	& 1.99	\tabularnewline
				Fixed-point \cite{li2013fixed}				&\textbf{2.13}      & 0.89  \tabularnewline
				\hline
				NN without ConvPP			&15.73		& 3.69		\tabularnewline
				ConvPP (ours)			&6.49		& \textbf{0.57}		\tabularnewline
				\hline 
				
			\end{tabular}

		}
		\hfill
		\parbox[t][][t]{.47\linewidth}{
			\centering
			\caption{Performance (error rate (\%)) of structured labeling methods on the FAQ sentence labeling dataset.}
			\label{res:FAQ}
			\begin{tabular}{l|c}
				\hline 
				Methods & error \tabularnewline
				\hline 
				Linear SVM \cite{do2010neural} & 9.87
				\tabularnewline
				Linear CRF \cite{vandermaaten2011hidden}				& 6.54	\tabularnewline
				NeuroCRFs \cite{do2010neural} & 6.05
				\tabularnewline
				Hidden-unit CRF \cite{vandermaaten2011hidden} 				& 4.43	\tabularnewline
				\hline
				NN without ConvPP 			& 5.25		\tabularnewline
				ConvPP (ours) 			& \textbf{1.09}		\tabularnewline
				\hline 
			\end{tabular}
		}
	\end{table}
	
	\textbf{Results.} 	The results in Table~\ref{res:OCR} and Table~\ref{res:FAQ} show that our proposed framework effectively models 1-D sequential structures and achieves better results for structured labeling as compared to previous methods. Several interesting observations: (1) on OCR dataset, compared to a kernel methods with hand-crafted features, our deep hybrid model performs worse on smaller dataset. But our deep learning approach performs better when the amount of training data increases. That is also the reason why ConvPP framework is important: handcrafted features and kernel methods are hard to be applied to many high-level vision tasks where big data is available. (2) ConvPP context window length reflects the range of context needed, we can see from Figure~\ref{fig:1d_curve} (a) that the generalization error converges when the context window length is about 7, which is the typical length of a word in the dataset. (3) Figure~\ref{fig:1d_curve} (b) shows that though we set the max number of testing iterations to be 10, with only 3 to 4 iterations at test time, the generalization error converges. That shows that the inference of our ConvPP model can be efficient. (4) The experiment in the simple sentence classification task shows that ConvPP has the potential to be applied on more NLP tasks such as sequence modeling.  (5) To show the effectiveness of the proposed approach, especially the convolutional pseudoprior part, we also perform the ablation study where we train a bottom-up network with exactly the same parameter settings. From the results we can see that without the structural information learned from the output label space, the performance decreases a lot.

	\subsection{Image Semantic Labeling: 2-D case}
	
		We then focus on two more challenging image labeling tasks: semantic segmentation and scene labeling. 
	Most of deep structured labeling approaches evaluate their performance on the popular Pascal VOC object segmentation dataset~\cite{pascal-voc-2012}, which contains only 20 object categories. Recently, CNN based methods, notably built on top of FCN \cite{long2015fully}, succeeded and dominated the Pascal-VOC leader-board where the performance (mean I/U) saturated to around 80\%. Here we instead evaluate our models on the much more challenging Pascal-Context dataset~\cite{mottaghi2014role}, which has 60 object/stuff categories and is considered as a fully labeled dataset (with much fewer pixels labeled as background). We believe the top-down contextual information should play a more crucial role in this case. We also evaluated our algorithm on SIFT Flow dataset~\cite{liu2009nonparametric} to evaluate our algorithm on the task of traditional scene labeling. In both experiments, the performance is measured by the standard mean intersection-over-union (mean I/U).
			
	\textbf{Multi-scale integration with FCN.} We build our hybrid model using FCN as the bottom-up CNN, and directly use the pre-trained models provided by the authors. FCN naturally handles multi-scale predictions by upgrading its 32-stride (32s) model to 16-stride (16s)/8-stride (8s) variants, where the final labeling decisions are made based on both high-level and low-lever representations  of the network. For the 32s model and an input image size $384 \times 512$, the size of the final output of FCN, after 5 pooling layers, is $12\times16$. As discussed in our formulation, ConvPP can be integrated into FCN by downsampling the ground-truth maps accordingly. 
	
%

\begin{figure}[t]
		\begin{center}
			\includegraphics[width=0.7\linewidth]{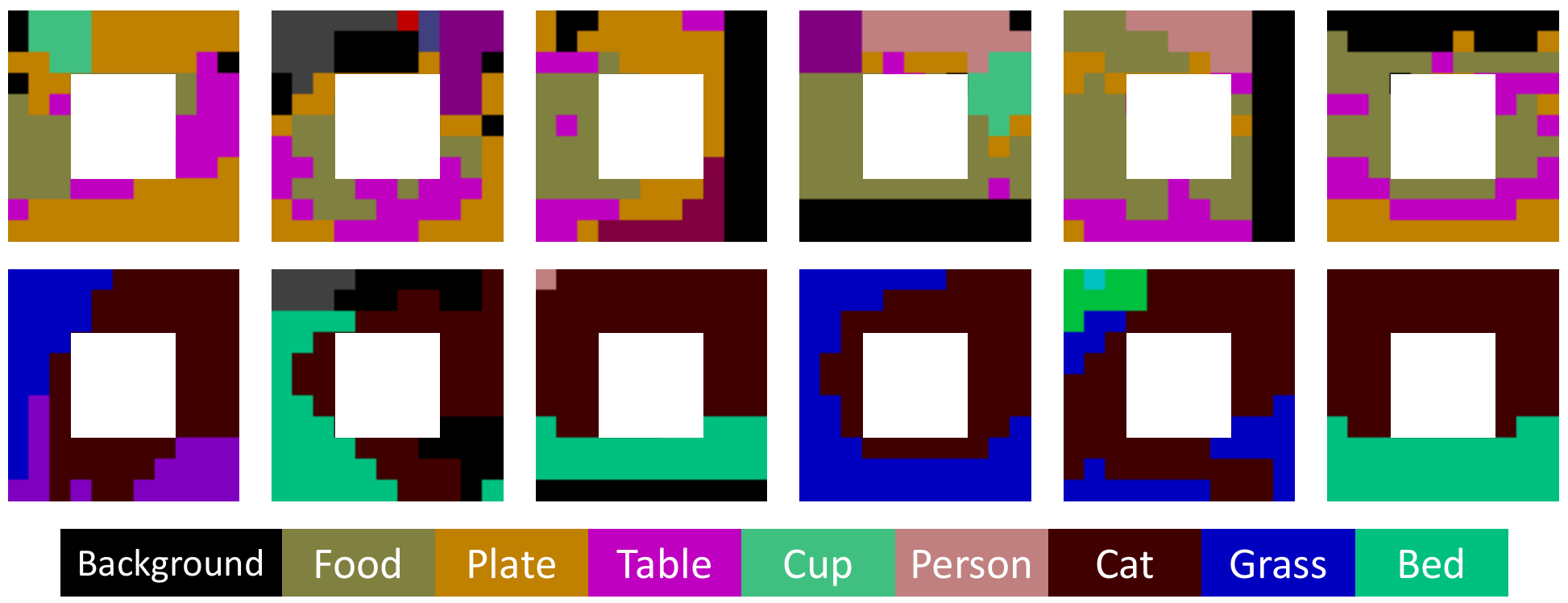} 
			\caption{ Visualization of $2$ filters. Each row displays top $6$ label patches that produce highest activation for a filter.}
			\label{fig:act}
		\end{center}
	\end{figure}
	
	\textbf{Hyper-parameter settings.} For all the 2-D labeling experiments, the number of channels in the donut filter convolution layer is 128. Adding more filters does not improve the performance. We keep all the hyper-parameters for the original bottom-up FCN network. The learning rates for the top-down ConvPP network are set to be 1e-7 for 32s, 1e-8 for 16s and 1e-9 for 8s variant. We choose the kernel size $k$ of our donut filters by cross validation. The size of the hole in the middle is set to $\lfloor k/2 \rfloor \times \lfloor k/2 \rfloor$. In following two image labeling experiments we evaluate two configurations of donut filter, namely donut filter with small $(7\times7)$ kernel size, and large $(11\times11)$ kernel size. The comparison of results for those two configurations is shown in Table~\ref{tab:kernelsize_results}. The choice of donut filter size is crucial to the pseudoprior representation learning.   
	\begin{wraptable}{r}{0.5\textwidth}
			\caption{\footnotesize Comparison of the results by varying the donut filter kernel size.}
			\label{tab:kernelsize_results}
			\newcommand{\tabincell}[2]{\begin{tabular}{@{}#1@{}}#2\end{tabular}}
			\centering
				
					\begin{tabular}{c|c|c}
						\hline 
						dataset & kernel size & mean IU    \tabularnewline
						\hline 
						PASCAL-Context & \tabincell{c}{$7\times7$ \\ $\mathbf{11\times11}$ } & \tabincell{c}{$40.3$ \\ {$\mathbf{41.0}$}}\\\hline
						SIFT Flow & \tabincell{c}{$\mathbf{7\times7}$ \\ $11\times11$ } & \tabincell{c}{$\mathbf{40.7}$ \\ $32.4$}\\\hline	
					\end{tabular}
		\end{wraptable}
		
	\textbf{Sparse donut filter.} We use $11\times11$ donut filters with $5\times5$ holes for Pascal-Context dataset since it achieves the best performance. The kernel covers a large portion of the 32-stride downsampled ground-truth map, and is therefore able to capture long range context. However, these large filters are typically very hard to learn. Inspired by \cite {chen2015semantic}, we reduce the number of learnable parameters while keeping the context range. Starting from a randomly initialized 6x6 kernel, we dilate the convolution kernel by inserting zeros between every neighboring position. Zero-valued locations are fixed to zero throughout the training. The resulting kernel is of size 11x11 but only 6x6 parameters are learnable.

	\textbf{Training and testing process.} 
	We follow the procedure of FCN to train our multi-scale hybrid model by stages. We train the ConvPP-32s model first, then upgrade it to the ConvPP-16s model, and finally to the ConvPP-8s model. During testing, we found that $3$ iterations are enough for our fixed-point approach to converge, thus we keep this parameter through out our experiments. 
	One concern is if the iterative testing process could diverge. Interestingly, in all our experiments (1-D and 2-D), the results are improved monotonically and converged. This shows that the pseudoprior learning process is stable and the fixed-point solver is effective.
	The input of ConvPP part is initialized with original FCN prediction since it is readily available. 
	
	\textbf{Computational cost.}
	Since we can utilize pretrained bottom-up network, training the single-layer top-down convolutional pseudoprior network is efficient. For Pascal-context dataset the training can be done in less than 1 hour on a single Tesla K40 GPU. The additional computational cost due to iterative inference procedure is also small. For 3 iterations of fixed-point inference, our ConvPP model only takes additional 150ms. Note that all previous works using CRFs (either online or offline) also require testing-stage iterative process.

		\begin{wraptable}{r}{0.5\textwidth}
		\caption{\footnotesize Results on Pascal-Context dataset~\cite{mottaghi2014role}. ConvPP outperforms FCN baselines and previous state-of-the-art models. $\dagger$ is trained with additional data from COCO.}
		\label{tab:pascalcontext_results}
		\begin{center}
				\begin{tabular}{l|c}
					\hline 
					&mean IU        \tabularnewline
					\hline 
					$\textrm{O}_{\textrm{{\tiny 2}}}\textrm{P}$ \cite{carreira2012semantic}             & 18.1    \tabularnewline
					CFM~(VGG+SS) \cite{dai2015convolutional}        & 31.5    \tabularnewline
					CFM~(VGG+MCG) \cite{dai2015convolutional}        & 34.4        \tabularnewline
					
					CRF-RNN \cite{zheng2015conditional}    &    39.3    \tabularnewline
					BoxSup$\dagger$  \cite{dai2015boxsup}    &    40.5    \tabularnewline
					\hline
					\hline
					FCN-32s \cite{long2015fully}             &    35.1    \tabularnewline
					ConvPP-32s    (ours)        &     37.1        \tabularnewline
					\hline 
					FCN-16s \cite{long2015fully}        &    37.6    \tabularnewline
					ConvPP-16s    (ours)        &     40.3        \tabularnewline
					\hline 
					FCN-8s \cite{long2015fully}            & 37.8        \tabularnewline
					ConvPP-8s    (ours)        &     \bf{41.0}    \tabularnewline
					\hline 
				\end{tabular}
		\end{center}
	\end{wraptable}	
	
	\textbf{Pascal-Context.} This dataset contains ground truth segmentations fully annotated with 60-category labels (including background), providing rich contextual information to be explored. We follow the standard training $+$ validation split as in ~\cite{mottaghi2014role,long2015fully}, resulting in 4,998 training images and 5,105 validation images. 
	
		
	Table \ref{tab:pascalcontext_results} shows the performance of our proposed structured labeling approach compared with FCN baselines and other state-of-the-art models. 

	We hope to evaluate our approach in a way that allows fair comparison with FCN, which does not explicitly handle structural information. Therefore we carefully control our experimental settings as follows: \textbf{(1)} We do not train the bottom-up CNN models for all the experiments in Training Stage 1, and use the pre-trained models provided by the authors.
	\textbf{(2)} We train the top-down ConvPP network (Training stage 2) independently on each scale, namely 32s, 16s and 8s.
	\textbf{(3)} To train the hybrid models at a certain scale, we only use the pre-trained FCN models at the corresponding scale. (ConvPP-32s can only use the FCN-32s representations.)
	\textbf{(4)} For all the experiments, we fix the learning rate of the FCN part of the hybrid model, namely all the convolutional layers from Conv1\_1 to fc7, to be zero. The reason we freeze the learning rate of FCN is to do an \textbf{ablation study}: we want to show the performance gain incorporating the ConvPP part. Intuitively context information should help more in high-level structural prediction rather than improving low-level appearance features. The experiment results support this claim: we get 40.89 when joint-tuning the parameters in the bottom-up FCN-8s network, the difference is negligible. Our methods consistently outperform the FCN baselines. We show the results for ConvPP 32s (structural information integrated in layer pool 5), ConvPP 16s (pool5+pool4) and ConvPP 8s (pool5 + pool4 + pool3) to analyze the effect of multi-scale context learning. The results have been consistently improved by combining finer scales.
	
	\begin{figure}[t]
		\center
		\includegraphics[width=0.9\linewidth]{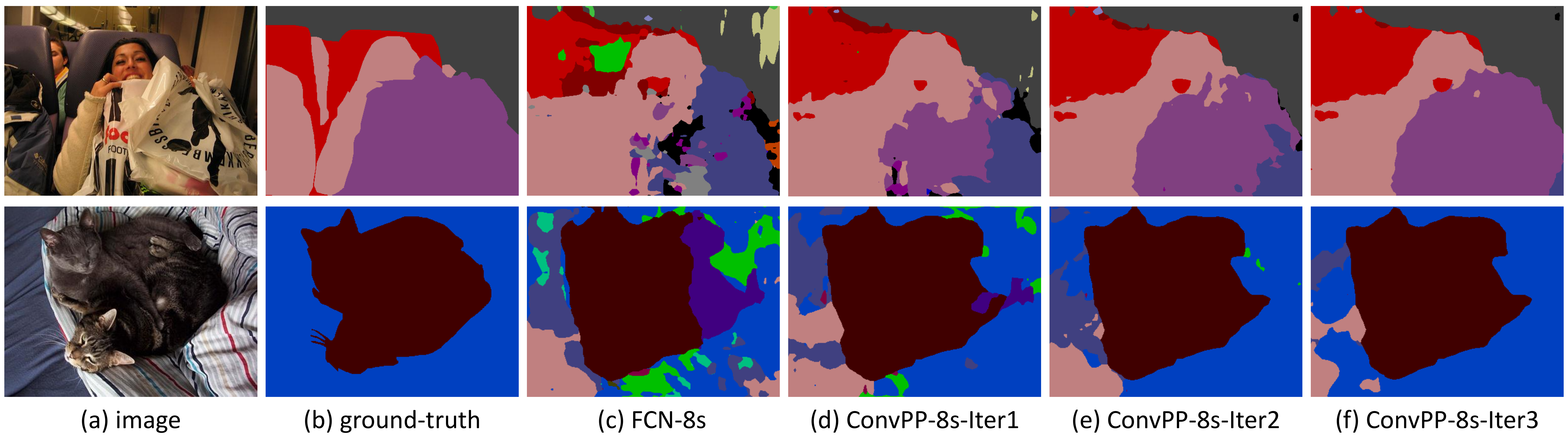} 
		\caption{\footnotesize Iterative update of labeling results during testing. Segmentation results are gradually refined. }
		\label{fig:qual_recur} 
		\par
	\end{figure}

		Our method also outperforms other state-of-the-art models built on FCN, notably CRF-RNN~\cite{zheng2015conditional}, which also explicitly handles structured labeling problem by integrating a fully-connected CRF model into the FCN framework; and BoxSup~\cite{dai2015boxsup}, which is trained with additional COCO data. This clearly shows the effectiveness of our ConvPP model in capturing complex inter-dependencies in structured output space. Example qualitative results of our ConvPP-8s compared to baseline FCN-8s model can be found in the supplementary material. Figure~\ref{fig:qual_recur} shows how our labeling results are iteratively refined at inference time. With multi-scale architecture design, our method leverages both short- and long-range context to assist the labeling task. ConvPP is able to recover correct labels as well as suppress erroneous label predictions based on contextual information. In Figure~\ref{fig:act}, we visualize $2$ learned donut filters on 32-stride ground-truth maps by displaying label patches that produce top $6$ activations (as done in \cite{zeiler2014visualizing}).  It is shown that our filters are learned to detect complex label context patterns, such as ``food-plate-tabel'', ``cat on the bed'' and ``cat on the grass''.

			
		
	\textbf{SIFT Flow} 
	We also evaluate our method on scene labeling task, where context is also important in accurate labeling. SIFT Flow dataset~\cite{liu2009nonparametric}  contains 2,688 images with 33 semantic categories. A particular challenge for our ConvPP model for this dataset is the relatively small image/ground-truth map size~($256\times256$), which means the 32-stride output is only $8\times8$. Downsampling the ground-truth map to this scale could potentially lead to loss in useful context information. In addition, the finest model provided by \cite{long2015fully} is FCN-16s instead of FCN-8s. 
	
		\begin{wraptable} {r}{0.53\textwidth}
			\caption{\footnotesize Results on SIFT Flow dataset \cite{liu2009nonparametric}.  Our methods outperform the strong FCN baselines. Improvement of ConvPP-8s vs FCN-8s is more significant than that of ConvPP-16s vs FCN-16s, since higher resolution ground truth map carries more structured information.}
			\label{tab:siftflow_results}
			\begin{center}
					\begin{tabular}{l|c}
						\hline 
						&mean IU        \tabularnewline\hline
						FCN-16s \cite{long2015fully}        &    39.1    \tabularnewline
						ConvPP-16s    (ours)        &     39.7        \tabularnewline
						\hline 
						FCN-8s \cite{long2015fully}            & 39.5        \tabularnewline		
						ConvPP-8s    (ours)        &     \bf{40.7}    \tabularnewline
						\hline 
						
					\end{tabular}
			\end{center}
		\end{wraptable}
	To alleviate this problem, we train our own FCN-8s model (pre-trained with the provided FCN-16s model) as our baseline and build our ConvPP-8s on top of it. Also because of the size of the image in the dataset, as shown in Table~\ref{tab:kernelsize_results}, $11\times11$ donut filters perform poorly. Thus we choose the donut filters with kernel size $7\times7$ and hole size $3\times3$, and the sparsification operation is not needed. The testing procedure is the same as that of Pascal-Context dataset.

		According to Table \ref{tab:siftflow_results}, our ConvPP models consistently outperform corresponding FCN baselines. The improvement of ConvPP-16s model is relatively small, which might result from the limited resolution of ground-truth maps (256 x 256). With higher ground-truth resolution, ConvPP-8s outperforms the stronger FCN-8s baseline by 1.2\% in mean I/U. This substantiate that our proposed pseudoprior learning framework is effective in learning structural information from the ground-truth labeling space.

%

	\section{Conclusions}
	We propose a new method for structured labeling by developing convolutional pseudoprior (ConvPP) on the ground-truth labels. ConvPP learns convolution in the labeling space with improved modeling capability and less manual specification. The automatically learns rich convolutional kernels can capture both short- and long- range contexts combined with a multi-scale hybrid model architecture design. We use a novel fixed-point network structure that facilitates an end-to-end learning process. Results on structured labeling tasks across different domains shows the effectiveness of our method. 
	\section* {\small{Acknowledgment}} This work is supported by NSF IIS-1618477, NSF IIS-1360566, NSF IIS-1360568, and a Northrop Grumman Contextual Robotics grant. We thank Zachary C. Lipton, Jameson Merkow, Long Jin for helping improve this manuscript. We are grateful for the generous donation of the GPUs by NVIDIA. 
	
	\bibliographystyle{splncs}
	\bibliography{0769}
	\clearpage
	\title{\footnotesize{Supplementary Material for Paper: \\Top-Down Learning for Structured Labeling with Convolutional Pseudoprior}} 
\author{}
\date{}

\maketitle

		\begin{figure}[!htp]
			\begin{centering}
				\vspace{-16mm}
				\includegraphics[width=0.8\linewidth]{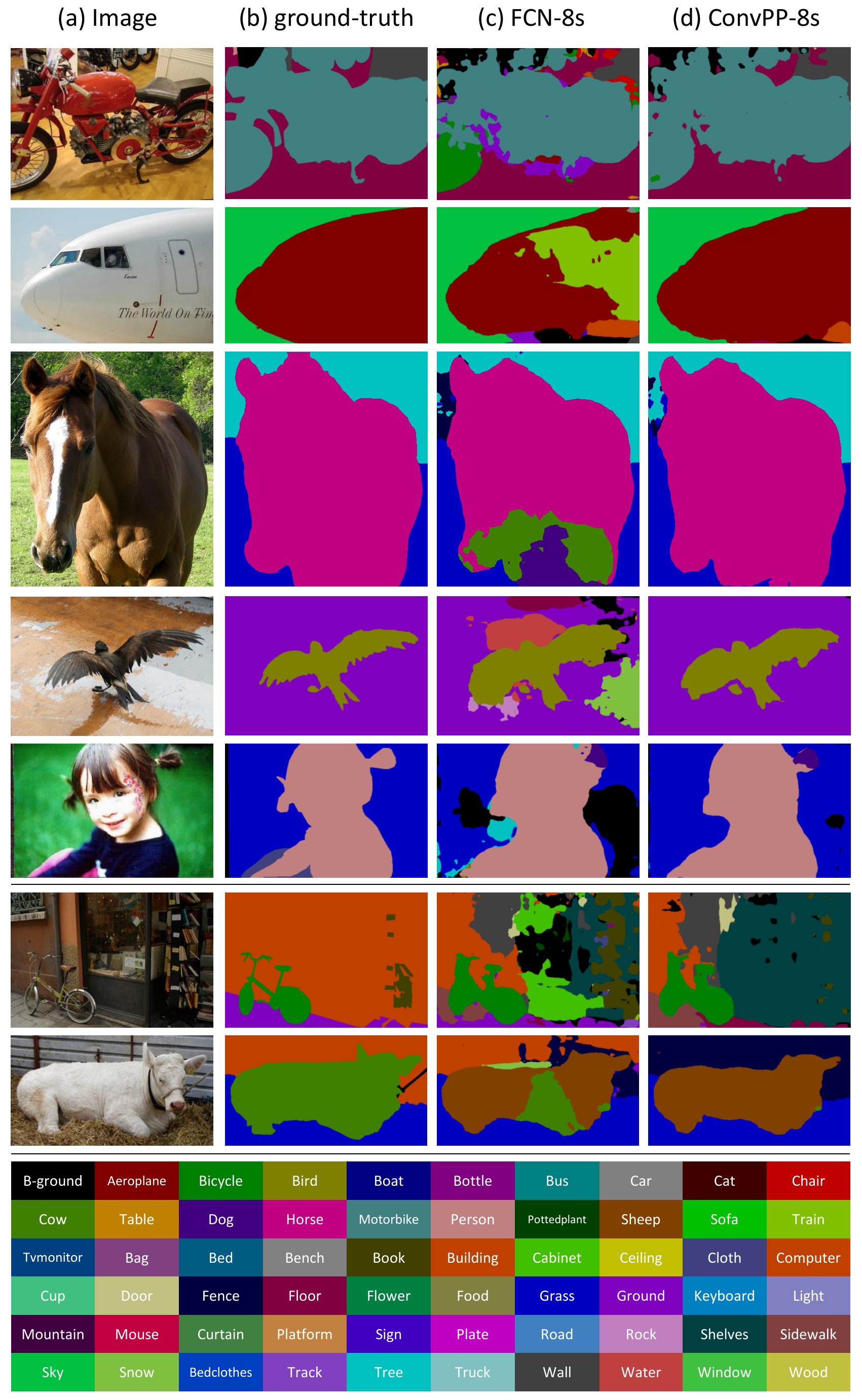} 
				\label{fig:qual3}
		\par\end{centering}
		\end{figure}
	
\end{document}